\documentclass[11pt]{article}

\usepackage[final]{acl}

\usepackage{times}
\usepackage{latexsym}
\usepackage{amsfonts}
\usepackage{enumitem}
\usepackage{amsmath}
\usepackage{amssymb}  % 提供♡、♣符号，必须早于author加载
\usepackage{booktabs}
\usepackage{multirow}
\usepackage{makecell}
\usepackage[most]{tcolorbox}
\usepackage{xcolor}
\usepackage[T1]{fontenc}
\usepackage[utf8]{inputenc}
\usepackage{microtype}
\usepackage{inconsolata}
\usepackage{graphicx}
\usepackage{array}
\usepackage{ragged2e}
\usepackage{listings}
\usepackage{footmisc}  % 优化脚注
\usepackage{wasysym}
\usepackage{fontawesome5}

\newcommand{\rsp}{RAGShaper}
\newcommand{\ic}{\texttt{InfoCurator}}

\newcommand{\email}[1]{\href{mailto:#1}{#1}}

\usepackage{hyperref}
\hypersetup{
    colorlinks=true,
    linkcolor=blue,
    urlcolor=blue,
    citecolor=blue
}

\usepackage{colortbl}
\usepackage{soul}
\usepackage{xcolor}

\newcommand{\ada}{$\textcircled{\raisebox{-.5pt}{1}}$}
\newcommand{\adb}{$\textcircled{\raisebox{-.5pt}{2}}$}

\usepackage{amsmath,amsfonts,bm}

\def\eqref#1{equation~\ref{#1}}
\def\1{\bm{1}}

\DeclareMathAlphabet{\mathsfit}{\encodingdefault}{\sfdefault}{m}{sl}
\SetMathAlphabet{\mathsfit}{bold}{\encodingdefault}{\sfdefault}{bx}{n}

\def\gA{{\mathcal{A}}}

\def\gQ{{\mathcal{Q}}}

\def\gT{{\mathcal{T}}}

\def\sD{{\mathbb{D}}}
\def\sK{{\mathbb{K}}}

\title{\rsp: Eliciting Sophisticated Agentic RAG Skills via \\ Automated Data Synthesis}

\author{
Zhengwei Tao\textsuperscript{\ada}\textsuperscript{\faPenNib}\thanks{Equal Contributions. \protect\faPenNib~Project Leads.},
Bo Li\textsuperscript{\ada}$^{*}$,
Jialong Wu\textsuperscript{\ada}\textsuperscript{\faPenNib},  
Guochen Yan\textsuperscript{\ada}, 
Huanyao Zhang\textsuperscript{\ada}\\
\textbf{
Jiahao Xu\textsuperscript{\adb},
Haitao Mi\textsuperscript{\adb},
Wentao Zhang\textsuperscript{\ada}\thanks{Corresponding Author.}
} \\
\textsuperscript{\ada}Peking University,
\textsuperscript{\adb}Tencent AI Lab
\\
\email{\{tttzw, wentao.zhang\}@pku.edu.cn}, \email{wujialongml@gmail.com}\\
}

\begin{document}
\maketitle
\begin{abstract}
Agentic Retrieval-Augmented Generation (RAG) empowers large language models to autonomously plan and retrieve information for complex problem-solving. However, the development of robust agents is hindered by the scarcity of high-quality training data that reflects the noise and complexity of real-world retrieval environments. Conventional manual annotation is unscalable and often fails to capture the dynamic reasoning strategies required to handle retrieval failures. To bridge this gap, we introduce \rsp, a novel data synthesis framework designed to automate the construction of RAG tasks and robust agent trajectories. \rsp~incorporates an \ic to build dense information trees enriched with adversarial distractors spanning \textit{Perception} and \textit{Cognition} levels. Furthermore, we propose a constrained navigation strategy that forces a teacher agent to confront these distractors, thereby eliciting trajectories that explicitly demonstrate error correction and noise rejection. Comprehensive experiments confirm that models trained on our synthesized corpus significantly outperform existing baselines, exhibiting superior robustness in noise-intensive and complex retrieval tasks.
\end{abstract}

\setlength{\abovedisplayskip}{3pt}
\setlength{\belowdisplayskip}{3pt}

\section{Introduction}
Agentic Retrieval-Augmented Generation (Agentic RAG) has emerged as a pivotal advancement in natural language processing, rapidly evolving from simple retrieval-and-read pipelines to autonomous systems capable of complex reasoning and dynamic tool usage~\cite{jin2025search, asai2024self, li2025search, team2025tongyi}. As Large Language Models (LLMs) are increasingly deployed in open-ended environments, Agentic RAG serves as the core infrastructure for a wide array of sophisticated applications, ranging from autonomous research assistants to domain-specific decision support systems. By endowing models with the agency to actively plan retrieval steps, evaluate gathered information, and iteratively refine their search, this paradigm represents a significant leap forward in bridging the gap between static knowledge bases and intelligent responses~\cite{singh2025agentic}.

\begin{figure}
  \centering
\includegraphics[width=\linewidth]{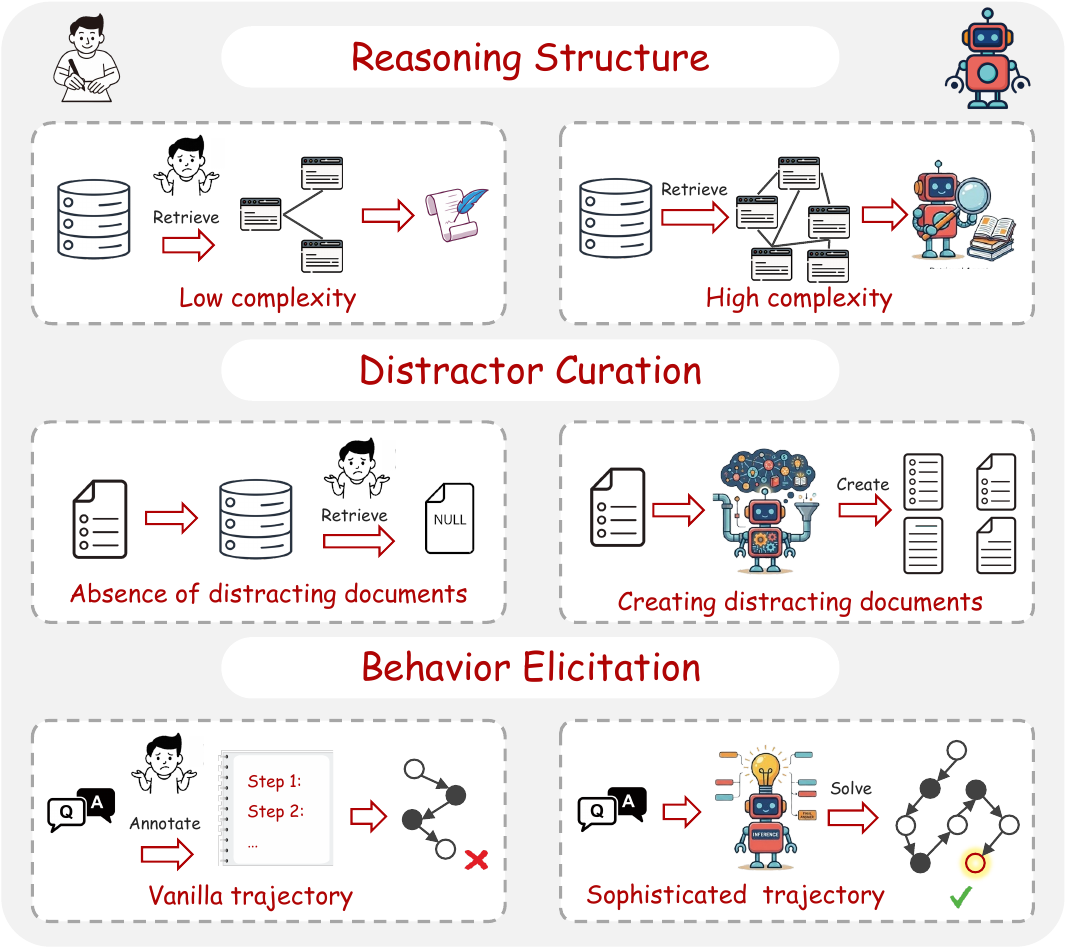}
   % \vspace{-3mm}
   \caption{Limitations of human annotation of the agentic RAG dataset, which can be mitigated by automatic synthesis by the agent curator.}
   \label{fig: intro}
   \vspace{-5mm}
\end{figure}

Current methodologies predominantly rely on manually annotated datasets, typically structured as question-trajectory-answer triplets~\cite{yang2018hotpotqa, ho2020constructing}. 
However, this paradigm is fundamentally ill-suited for training Agentic RAG models due to the intrinsic cognitive and operational bottlenecks of human annotators, as shown in Figure~\ref{fig: intro}. 
\textbf{First}, constrained by limited working memory, annotators struggle to synthesize implicit, multi-hop evidence scattered across a large volume of disparate documents, often defaulting to shallow, single-context reasoning rather than the deep retrieval chains required for robust agents~\cite{wu2025webdancer}. 
\textbf{Second}, manually curating realistic, noise-heavy retrieval environments is impractical. 
Retrieval distractors that are lexically similar yet factually incorrect may not exist~\cite{yan2024corrective}. 
\textbf{Finally}, human annotations is hard to capture the dynamic strategy adjustments required to decompose tasks and recover from retrieval failures~\cite{jeong2024adaptive,tian2025right}. Consequently, these limitations make the manual construction of high-quality data for Agentic RAG difficult to scale. 

To surmount these impediments and automate the production of high-fidelity training corpora, we introduce \textbf{\rsp}, a novel framework specifically engineered for Agentic RAG data synthesis. Addressing the complexity of information construction, \rsp~incorporates an \texttt{InfoCurator} module designed to autonomously build a comprehensive retrieval environment. Starting from a seed entity, the curator leverages retrieval tools to perform multi-round exploration within the knowledge base, aggregating a dense information tree of entities and interrelations to support the synthesis of tasks requiring deep reasoning. Beyond gathering positive evidence, the curator dynamically generates adversarial ``distractor'' documents based on the retrieved context. 
We systematically categorize these distractors into two dimensions, \textit{Perception} and \textit{Cognition}, a taxonomy designed to cultivate robust agentic discrimination capabilities against varying levels of information noise. Following information curation, an LLM utilizes this structured context to synthesize specific tasks and corresponding ground-truth answers. To extract optimal skill and behavior patterns, we employ a sophisticated teacher agent to solve these synthesized tasks; uniquely, we enforce a constrained navigation strategy that mandates the retrieval of the generated distractors, thereby explicitly capturing the teacher's adaptive strategies in identifying and overcoming information hazards. Finally, by fine-tuning a base model on this large-scale corpus of agent trajectories, we obtain a robust Agentic RAG model proficient in navigating noisy environments.

We summarize our contributions as follows:
\begin{itemize}[left=0.2cm]\setlength{\itemsep}{0pt}\setlength{\parskip}{0pt}\setlength{\topsep}{-5pt}
    \item We introduce \rsp, an agentic RAG data synthesis framework featuring an \texttt{InfoCurator} designed to aggregate densely connected information and synthesize sophisticated retrieval distractors across multiple dimensions.
    \item We propose a constrained navigation strategy to elicit robust error-correction and reasoning behaviors from the teacher agent, enabling the large-scale accumulation of high-quality, resilient trajectories.
    \item We conduct extensive experiments to validate our data synthesis framework, with empirical results demonstrating that models trained on our corpus significantly outperform baselines in complex retrieval environments.
\end{itemize}
% \vspace{-5mm}
\section{Preliminaries}
\label{preliminaries}
% \vspace{-2mm}

We formalize the Agentic RAG framework as an autonomous agent that interleaves reasoning with retrieval, enabling to dynamically interact with external corpora to resolve knowledge-intensive queries. Adopting the ReAct paradigm~\citep{yao2023react}, the agent navigates a sequential decision process where it must iteratively bridge the gap between its internal knowledge and the required external evidence. At each time step $t$, the agent conditions on the initial query and the history of prior interactions to generate a reasoning thought $\tau_t$. This reasoning guides the selection of a specific retrieval tool-use action $\alpha_t$, such as querying a knowledge base $\sK$ to retrieve documents $\sD$, which yields a corresponding observation $o_t$.
This cumulative reasoning-retrieval loop is represented by the \emph{agent trajectory}, denoted as:
\begin{equation}
\label{eq: trajectory}
\gT = (\gQ, \tau_1, \alpha_1, o_1, \ldots, \tau_{T}, \alpha_{T}, o_{T}, \gA),
\end{equation}
where $\gQ$ represents the user task, and the tuple $(\tau_i, \alpha_i, o_i)$ captures the agent's planning, tool-use action, and feedback at step $i$. $\gA$ denotes the final answer for $\gQ$, representing the agent's primary objective. The purpose of our data synthesis is to construct $(\gQ, \gA, \gT)$ triples for RAG agent training.

\begin{figure*}
  \centering
  \setlength{\belowcaptionskip}{-4mm}
\includegraphics[width=0.89\linewidth]{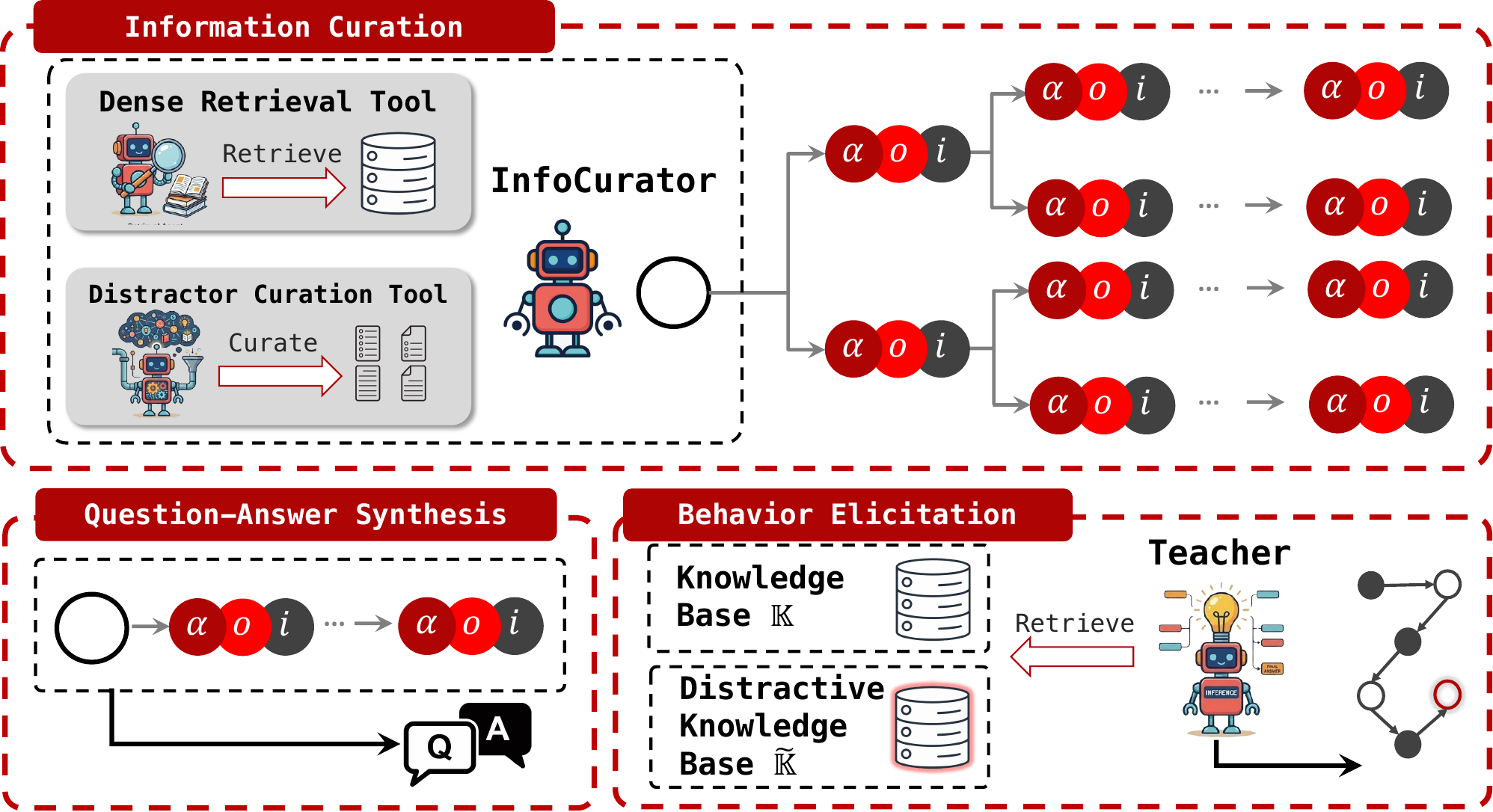}
   % \vspace{-3mm}
   \caption{Overview of RAGShaper.}
   \label{fig: overview}
   % \vspace{-5mm}
\end{figure*}

\vspace{-2mm}
\section{Method}
\label{sec: method}
\vspace{-2mm}

We propose \rsp, a data synthesis framework designed to automate the construction of high-quality training corpora for Agentic RAG. As illustrated in Figure~\ref{fig: overview}, our pipeline consists of four phases: (1) \textbf{Information Curation} (\S\ref{sec:information_curation}), where an autonomous curator agent explores a seed entity to build a dense, distractor-augmented information tree, followed by a selection process to identify useful information paths; (2) \textbf{Question-Answer Synthesis} (\S\ref{sec:qa_synthesis}), where tasks are derived from these selected paths; (3) \textbf{Behavior Elicitation} (\S\ref{sec:bahavior}), where a teacher agent solves these tasks under a specific distraction strategy to generate trajectories exhibiting sophisticated behaviors; and (4) \textbf{Training} (\S\ref{sec:training}), where the student model is fine-tuned on these enhanced trajectories.

\subsection{Information Curation}
\label{sec:information_curation}
To train agents capable of deep reasoning, the underlying information retrieval tasks must be rich in inter-entity relationships and semantically challenging noise. As manually constructing such information structures is not scalable, we design an \ic~ agent to automate this process.

\subsubsection{Tree Exploration on \ic}
The goal of the \ic~ is to construct an information structure from a knowledge base $\sK$, which serves as the foundation for subsequent question synthesis. Specifically, \ic~ builds an information tree by retrieving positive facts and crafting distractive documents.
The exploration begins with a seed entity, which serves as the root node $s_1$ of the tree. \ic~ then expands new nodes via depth-first traversal to explore new information. A node is defined as:
\begin{equation}
s_t =
\begin{cases}
\text{seed entity}, & t = 1 \\
\{\alpha_t,\ i_t,\ o_t\},      & t > 1,
\end{cases}
\end{equation}
where $\alpha_t$ and $i_t$ represent the action and intention of \ic~ for expanding the node. The action $\alpha_t$ involves either retrieving documents or creating distractive ones (detailed below), and $o_t$ is the observation resulting from $\alpha_t$. We crawl large-scale entities from Wikipedia\footnote{https://www.wikipedia.org/}.
The agent expands a node by invoking tools based on the path from the current node to the root:
\begin{equation}
\begin{aligned}
        \alpha_{t+1}, i_{t+1} & = \ic ( \text{Path} (s_1, s_t)), \\
    o_{t+1} & = \texttt{Execute}(\alpha_t). 
\end{aligned}
\end{equation}

\begin{table*}[!t]
\tiny
  \setlength{\belowcaptionskip}{-2mm}
    \centering
    \resizebox{\linewidth}{!}{
    \begin{tabular}{>{\RaggedRight}p{1.2cm} >{\RaggedRight}p{1.8cm} >{\RaggedRight}p{2.8cm} >{\RaggedRight}p{3.2cm} >{\RaggedRight}p{2.5cm}}
        \toprule
        \textbf{Level} & \textbf{Type} & \textbf{Description} & \textbf{Example} & \textbf{Target Agent Skill} \\
        \midrule
        \multirow{3}{*}{\makecell[l]{\textit{Perception}\\\textit{Layer}}} 
        & \multirow{3}{*}{Doppelgänger}  
        & Contains core topics of the query but with different metadata (version/date/ID). 
        & Question: \textcolor{blue}{\textbf{2024}} Financial Report. \newline Distractor: \textcolor{red}{\textbf{2025}} Financial Report. 
        & \textit{\textbf{Precision Verification}}: Verify metadata to avoid being misled by similarity. \\
        \midrule
        \multirow{11}{*}{
        \makecell[l]{\textit{Cognition}\\\textit{Layer}}
        % Cognition Layer
        } 
        & \multirow{3}{*}{False Shortcut} 
        & Forged A→C direct connection (real logic: A→B→C) with ambiguous/wrong justifications. 
        & Truth: Virus $\to$ \textcolor{blue}{\textbf{Fever}} $\to$ Weakness. \newline Distractor:
        ``Whether the virus causes weakness remains \textcolor{red}{\textbf{unknown}}''. 
        &  \textit{\textbf{Reasoning Persistence}}: Reject shortcuts; search for intermediate nodes. \\
        \cmidrule(lr){2-5}
        & \multirow{4}{*}{Fragmented Puzzle} 
        & The answer is distributed across several documents. 
        & Question: \textcolor{blue}{\textbf{How many}} years has the company been profitable? \newline  Distractor: Each distractor document includes content for \textcolor{red}{\textbf{a single year}}. 
        &  \textit{\textbf{Completeness Awareness}}: Identify information truncation; perform complete retrieval. \\
        \cmidrule(lr){2-5}
        & \multirow{4}{*}{Subjective Fallacy}  
        & Subjective tone with objectively wrong core arguments. 
        & Truth: Drug X \textcolor{blue}{\textbf{effectiveness is 95\%}}. \newline Distractor: Despite claims, \textcolor{red}{\textbf{I feel}} Drug X is useless.
        &  \textit{\textbf{Fact-Opinion Separation}}: Distinguish opinions from facts; reject unsupported views. \\
        \bottomrule
    \end{tabular}
    }
        \caption{Distractor types, examples, and corresponding target agent skills.}
    \label{tab: distractor_types}
\end{table*}

At each step, we expand two child nodes with probability $p^{b}$ and one node with probability $1-p^{b}$. Expansion terminates when the node depth reaches a predefined threshold. The resulting information tree contains facts and their relations. This automated process significantly alleviates the workload of manual data organization. We detail the tools used by \ic~ below.

\noindent\paragraph{Dense Retrieval Tool.}
\ic~ is equipped with a Dense Retrieval Tool. The parameters include a \textit{Query} and \textit{Topk}, representing the search string and the desired number of relevant documents, respectively. The tool encodes the query using a pretrained text embedding\footnote{We use E5 as the retriever in the DPR project: https://github.com/facebookresearch/DPR} and computes the similarity between the query and documents indexed in the KB. It returns documents with similarity scores exceeding a threshold $\tau$, ensuring the output count does not exceed $k$:
\begin{equation}
\label{eq: retrieval}
\begin{aligned}
    \sD & = R(\sK, k) \\
    & = \operatorname{Topk}\left( \{ d \in \sK \mid \operatorname{sim}(Query, d) > \tau \}\right),
\end{aligned}
\end{equation}
where $d$ represents a document and $\tau$ denotes the similarity threshold. 

\noindent\paragraph{Distractor Curation Tool.}
A robust Agentic RAG model must distinguish between relevant evidence and noise. Merely including positive facts in the information set is insufficient; we must also include challenging distractors. However, relying solely on retrieving similar facts from the KB as distractors is often impractical due to lack of precision or the absence of suitable candidates. Therefore, we introduce the Distractor Curation Tool, which directly generates and stores distractive documents. 

A distractive document is not necessarily factually incorrect but is designed to be confusing within the context of the RAG task. We include four types of distractors spanning both \textit{Perception} and \textit{Cognition} levels, as shown in Table~\ref{tab: distractor_types}. The tool takes an \textit{Original fact}, a \textit{Distractor type}, and a \textit{Creating guideline} as input, calling an LLM to generate a distracting fact based on these parameters. The guideline ensures the generated content is precise.

\subsubsection{Information Path Selection}
After building the information structure, we identify specific sub-structures for QA synthesis. The raw tree contains numerous divergent paths, not all of which form coherent reasoning chains. We employ a heuristic selection mechanism to extract high-value paths from the root to the leaves. We posit that a desirable path contains a high density of information. Thus, we score each path based on the total number of documents it contains, including both positive entries and distractors:
\begin{equation}
    \text{score}_l = \sum_{s \in \text{Path}(s_1, s_l)} |\mathcal{D}_s|,
\end{equation}
where $|\mathcal{D}_s|$ denotes the document count at node $s$, and $s_l$ is a leaf node. We select the $m$ paths with the highest scores for synthesis.

\subsection{Question-Answer Synthesis}
\label{sec:qa_synthesis}
Once the paths are selected, we synthesize the task $(\mathcal{Q}, \mathcal{A})$. Motivated by the need to align the question with the retrieval steps, we prompt an LLM to "reverse-engineer" a question that strictly requires the information sequence found in the path to answer. The generator conditions on the full sequence of observations and intents:
\begin{equation}
    (o^c_{1}, a^c_{1}, i_{1}, \ldots, o^c_{N}, a^c_{N}, i_{N}) \implies (\mathcal{Q}, \mathcal{A})
\end{equation}
Here, the inclusion of the intent $i$ is critical. By explicitly exposing the \ic's intent, the LLM ensures that $\mathcal{Q}$ naturally necessitates that specific information, guaranteeing that the path serves as a valid reasoning support.

\subsection{Behavior Elicitation}
\label{sec:bahavior}
After harvesting the $(\gQ, \gA)$ pairs, we construct the agent execution trajectory $\gT$. Directly using the curated path is suboptimal, as it may be noisy or not represent the most efficient solution. Instead, we employ a \texttt{Teacher} agent to solve $\gQ$, thereby generating the final training trajectory $\gT$:
\begin{equation}
    \Tilde{\gA}, \gT = \texttt{Teacher}(\gQ),
\end{equation}
where $\Tilde{\gA}$ is the predicted answer. The trajectory follows the format defined in Eq. (\ref{eq: trajectory}). The teacher agent is equipped only with the Dense Retrieval Tool, identical to \ic.

To elicit the sophisticated behaviors and abilities outlined in Table~\ref{tab: distractor_types}, we implement a specific strategy using the generated distractors. We aggregate all distractive documents into a secondary KB, $\Tilde{\sK}$. During retrieval, the tool fetches documents $\mathbb{D}_t$ from both the original KB and $\Tilde{\sK}$ according to the following logic:
\begin{equation}
    \left\{ 
    \begin{array}{@{}l@{\,\,}l} 
        R(\mathbb{K}, k-2) \cup R(\tilde{\mathbb{K}}, 2), & \text{if } t=1, \\[5pt]
        R(\mathbb{K}, k), & \text{if } \tilde{\mathbb{K}} \cap \mathbb{D}_{t-1} \neq \emptyset, \\[5pt]
        R(\mathbb{K}, k-2) \cup R(\tilde{\mathbb{K}}, 2), & \text{with prob. } p^e, \\[5pt]
        R(\mathbb{K}, k), & \text{otherwise}.
    \end{array}
    \right.
\end{equation}
where $p^e$ is a fixed probability. $R(\sK, k)$ is the retreival function defined in Eq. (\ref{eq: retrieval}). At the first step, the agent is forced to retrieve from $\Tilde{\sK}$. If retrieval from $\Tilde{\sK}$ occurred in the previous step, it is suppressed in the current step to prevent continuous hallucination loops. Otherwise, retrieval from $\Tilde{\sK}$ occurs with probability $p^e$. Crucially, the \texttt{Teacher} agent remains agnostic to the existence of $\Tilde{\sK}$. 

% \vspace{-2mm}
\subsection{Training}
\label{sec:training}
% \vspace{-2mm}

Finally, we compile the synthesized triples $(\gQ, \gA, \gT)$ into a training dataset, retaining only trajectories where the predicted answer is correct (i.e., $\Tilde{\gA}=\gA$). We follow common RAG evaluation by using F1 score to filter training data~\cite{jin2025search}. We only remain data where the F1 score is above 0.9. We fine-tune a base LLM to minimize the standard negative log-likelihood loss on the agent trajectory tokens, following standard supervised fine-tuning (SFT) protocols:
\begin{equation}
\begin{aligned}
L = 
- \frac{1}{\sum_{i=1}^{|\gT|} \mathbb{I}[x_i \ne o]}
\sum_{i=1}^{|\gT|} 
\mathbb{I}[x_i \ne o] \cdot \\
\log \pi_{\theta}(x_i \mid x_{<i}).
\end{aligned}
\end{equation}
where $x_i$ is the $i^{th}$ token and $\mathbb{I}$ is the indicator function masking observation tokens. By training on trajectories $\gT$ that include behaviors such as self-correction and distractor rejection, derived from our constrained elicitation process, the resulting model learns to operate autonomously in noisy, open-ended retrieval environments.

\begin{table*}
\centering
\small
% \resizebox{1\textwidth}{!}{%
\begin{tabular}{l|cccccccccc}
\toprule
\multirow{2}{*}{\textbf{Models}} & \multicolumn{2}{c|}{\textbf{Bamboogle}} & \multicolumn{2}{c|}{\textbf{PopQA}} & \multicolumn{2}{c|}{\textbf{NQ}} & \multicolumn{2}{c|}{\textbf{AmbigQA}} & \multicolumn{2}{c}{\textbf{Avg}} \\
\cmidrule{2-11}
 & \multicolumn{1}{c}{\textbf{EM}} & \multicolumn{1}{c|}{\textbf{F1}} & \multicolumn{1}{c}{\textbf{EM}} & \multicolumn{1}{c|}{\textbf{F1}} & \multicolumn{1}{c}{\textbf{EM}} & \multicolumn{1}{c|}{\textbf{F1}} & \multicolumn{1}{c}{\textbf{EM}} & \multicolumn{1}{c|}{\textbf{F1}} & \multicolumn{1}{c}{\textbf{EM}} & \multicolumn{1}{c}{\textbf{F1}} \\
\midrule
\multicolumn{11}{c}{\cellcolor{gray!10}\textit{\textbf{Prompt-Based Methods}}} \\
\midrule

Iter-RetGen & 14.4 & 23.9 & 42.5 & 49.3 & 34.5 & 44.2 & 47.0 & 58.8 & 34.6 & 44.1 \\
Selective-Context & 15.3 & 22.6 & 34.9 & 41.5 & - & - & - & - & - & - \\
LongLLMLingua & 20.3 & 27.4 & 39.2 & 45.1 & - & - & - & - & - & - \\
IR-COT & 16.0 & 27.9 & 32.4 & 39.9 & 19.3 & 35.5 & 24.5 & 40.6 & 23.1 & 36.0 \\
RECOMP & 21.7 & 28.6 & 40.5 & 45.8 & - & - & - & - & - & - \\
FLARE & 15.2 & 24.6 & 36.8 & 44.9 & 28.9 & 43.2 & 40.6 & 50.1 & 30.4 & 40.7 \\
Search-o1 & 30.4 & 39.9 & 47.0 & 50.0 & 30.3 & 40.7 & 42.5 & 53.4 & 37.6 & 46.0 \\

\midrule
\multicolumn{11}{c}{\cellcolor{gray!10}\textit{\textbf{Learning-Based Methods}}} \\
\midrule

Search-R1 & 30.4 & 43.2 & 41.3 & 46.4 & 36.0 & 45.0 & 49.2 & 60.4 & 39.2 & 48.8 \\
IKEA & 30.4 & 45.3 & 38.7 & 42.7 & 30.7 & 42.8 & 47.0 & 57.9 & 36.7 & 47.2 \\
ReasonRAG & 22.4 & 29.1 & 41.1 & 44.4 & 28.1 & 38.9 & 39.7 & 51.9 & 32.8 & 41.1 \\
DeepRAG & - & - & 40.6 & 43.2 & - & - & - & - & - & - \\
DecEx-RAG & 37.6 & 49.3 & \textbf{51.3} & \textbf{53.2} & 36.0 & 47.2 & 49.5 & 59.5 & 43.6 & 52.3 \\
HL-Data 4.5k & 50.4 & 67.5 & 35.2 & 48.3 & 31.5 & 47.4 & 52.1 & 69.0 & 42.3 & 58.0\\
\midrule
\multicolumn{11}{c}{\cellcolor{blue!20}\textit{\textbf{Ours}}} \\
\midrule
RAGShaper 4.5k & \underline{58.5} & \underline{70.3} & 37.4 & 47.8 & \underline{38.3} & \underline{50.0} & \textbf{61.3} & \textbf{71.4} & \underline{48.8} & \underline{59.8} \\
RAGShaper 6.5k & \textbf{60.0} & \textbf{72.6} & \underline{38.9} & \underline{49.6} & \textbf{41.3} & \textbf{54.8} & \underline{61.1} & \underline{71.1} & \textbf{50.3} & \textbf{62.0} \\
\midrule
\end{tabular}%
% }
\caption{Performance comparison on evaluation datasets. HL-Data denotes open-sourced human labeled data, i.e. sampled HotpotQA and 2WikiMultiHopQA from training set. Avg is recalculated based on Bamboogle, PopQA, NQ, and AmbigQA. \textbf{Bold} stands for the highest score, and \underline{underline} is the second best.}
\label{tab: main_results}
\end{table*}

\section{Experiments}

\subsection{Experimental Settings}
\label{sec: experimental_settings}

\noindent\paragraph{Data Synthesis.} We set the branch probability $p^{b}$ to 0.5 if it's on the first 2 depth of the exploration tree, otherwise $p^{b}=0$. The maximum depth of the tree is 30. The dense retrieval tool threshold $\tau$ is 0.8. The distractive probability $p^{e}$ in Behaviour Elicitation is 0.5. We select two paths ($m=2$) for data synthesis from each exploration tree. We use gpt-oss-120b as the Teacher agent, where \ic~is based on it as well.

\noindent\paragraph{Training.} We train on Qwen3-30B-A3B-Think and Qwen3-4B-Think~\cite{qwen3technicalreport} on Megatron-LM framework. We use 4.5k and 6k data settings. Details are in the Appendix~\ref{app: training_details}.

\noindent\paragraph{Evaluation Benchmark.} To comprehensively evaluate the reasoning and retrieval capabilities of our agent, we conduct experiments on four diverse open-domain RAG benchmarks: Natural Questions (NQ)~\cite{kwiatkowski2019natural}, PopQA~\cite{mallen2023popqa}, AmbigQA~\cite{min2020ambigqa}, and Bamboogle~\cite{press2023measuring}. We report performance using standard Exact Match (EM) and F1 Score metrics. We use evaluation setting the same as DecEx-RAG. Details are in Appendix~\ref{app: eval_benchmark}. 

\noindent\paragraph{Baselines.}We compare \rsp~against a wide range of competitive baselines. For prompt-based methods, we include Iter-RetGen~\cite{shao2023enhancing}, IR-CoT~\cite{trivedi2023interleaving}, FLARE~\cite{jiang2023active}, Selective-Context~\cite{li2023unlocking}, LongLLMLingua~\cite{jiang2024longllmlingua}, RECOMP~\cite{xu2023recomp}, and Search-o1~\cite{li2025search}. Regarding learning-based methods, we benchmark against DeepRAG~\cite{guan2025deeprag}, IKEA~\cite{huang2025reinforced}, ReasonRAG~\cite{zhang2025process}, DecEx-RAG~\cite{leng2025decex}, Search-R1~\cite{jin2025search}, and HL-Data~\cite{jin2025search, leng2025decex} (i.e. Subset of HotPotQA and 2Wiki, therefore we don't take them for evaluation). Detailed descriptions are provided in Appendix~\ref{app: baseline_details}.

% \vspace{-1mm}
\subsection{Main Results}
\label{sec: main_results}
% \vspace{-1mm}

\noindent\textbf{\rsp~establishes significant improvement.} Table~\ref{tab: main_results} presents the comparison of \rsp~against state-of-the-art baselines. Our method consistently achieves the best performance, with the 6.5k model setting a new benchmark of 50.3 Avg EM and 62.0 Avg F1, significantly surpassing both prompt-based (e.g., Search-o1) and learning-based methods.

\noindent\textbf{Synthesized data surpasses human annotation in quality.} Crucially, \rsp~demonstrates superior data efficiency compared to human annotation. Under the same data scale (4.5k), our method outperforms HL-Data across almost all metrics. This indicates that our automated pipeline generates higher-quality training data which excels traditional crowd-sourced data.

\noindent\textbf{Distractor training enables robustness on complex, noisy tasks.} The performance gains are most pronounced on complex, noise-intensive tasks like Bamboogle and AmbigQA. The significant lead on AmbigQA directly validates the effectiveness of our \textit{Distractor Curation} mechanism and \textit{Behaviour Elicitation}. By training on trajectories laden with multi-dimension of distractors, our agent effectively learns to filter retrieval noise and execute robust multi-hop reasoning, a capability essential for navigating the ambiguity inherent and adapting retrieving strategy in these challenging datasets.

% \vspace{-1mm}
\subsection{Ablation Study}
\label{sec: ablation}
% \vspace{-1mm}

To assess the contribution of our distractor-based learning mechanism, we conduct an ablation study using a variant named RAGShaper-Dis. We exclude the Distractor Curation Tool during data synthesis and remove the noise-injection strategy during the Behavior Elicitation phase. The agent is trained solely on clean, positive reasoning paths without exposure to adversarial retrieval contexts.

\noindent\textbf{Distractor-based learning is essential for robust retrieval.} As shown in Table~\ref{tab: ablation_results}, removing these components leads to a severe performance drop, with the Average EM plummeting from 48.8 to 33.8. The decline is most dramatic on noise-sensitive datasets like AmbigQA and Bamboogle. These results strongly underscore the necessity of our approach: training on "clean" data alone is insufficient for robust agentic retrieval. The proposed synthesis of perception and cognition-level distractors is essential for equipping the agent with the critical ability to discern evidence from noise in complex real-world environments.

\begin{table*}
\centering
\small
\setlength{\belowcaptionskip}{-3mm}
% \resizebox{1\textwidth}{!}{%
\begin{tabular}{l|cccccccccc}
\toprule
\multirow{2}{*}{\textbf{Models}} & \multicolumn{2}{c|}{\textbf{Bamboogle}} & \multicolumn{2}{c|}{\textbf{PopQA}} & \multicolumn{2}{c|}{\textbf{NQ}} & \multicolumn{2}{c|}{\textbf{AmbigQA}} & \multicolumn{2}{c}{\textbf{Avg}} \\
\cmidrule{2-11}
 & \multicolumn{1}{c}{\textbf{EM}} & \multicolumn{1}{c|}{\textbf{F1}} & \multicolumn{1}{c}{\textbf{EM}} & \multicolumn{1}{c|}{\textbf{F1}} & \multicolumn{1}{c}{\textbf{EM}} & \multicolumn{1}{c|}{\textbf{F1}} & \multicolumn{1}{c}{\textbf{EM}} & \multicolumn{1}{c|}{\textbf{F1}} & \multicolumn{1}{c}{\textbf{EM}} & \multicolumn{1}{c}{\textbf{F1}} \\
\midrule
\multicolumn{11}{l}{\textit{Qwen3-30B-A3B-Think}} \\
\midrule
RAGShaper-Dis 4.5k & 38.4 & 58.9 & 27.9 & 42.4 & 28.0 & 44.2 & 41.0 & 61.2 & 33.8 & 51.6 \\
\textbf{RAGShaper 4.5k} & 58.5 & 70.3 & 37.4 & 47.8 & 38.3 & 50.0 & 61.3 & 71.4 & 48.8 & 59.8 \\

\midrule
\multicolumn{11}{l}{\textit{Qwen3-4B-Think}} \\
\midrule

HL-Data 4.5k & 40.8 & 55.3 & 27.0 & 41.8 & 33.5 & 46.8 & 52.9 & 65.6 & 38.5 & 52.4 \\
\textbf{RAGShaper 4.5k} & 54.4 & 63.9 & 32.7 & 45.4 & 33.1 & 45.0 & 56.0 & 65.5 & 44.0 & 54.9 \\
\bottomrule
\end{tabular}%
% }
\caption{Ablation studies and experiments on different backbones. RAGShaper-Dis stands for experiments on distractive documents created and added in the Behaviour Elicitation.}
\label{tab: ablation_results}
\end{table*}

\begin{figure}

  \centering
  \setlength{\belowcaptionskip}{-3mm}
\includegraphics[width=0.78\linewidth]{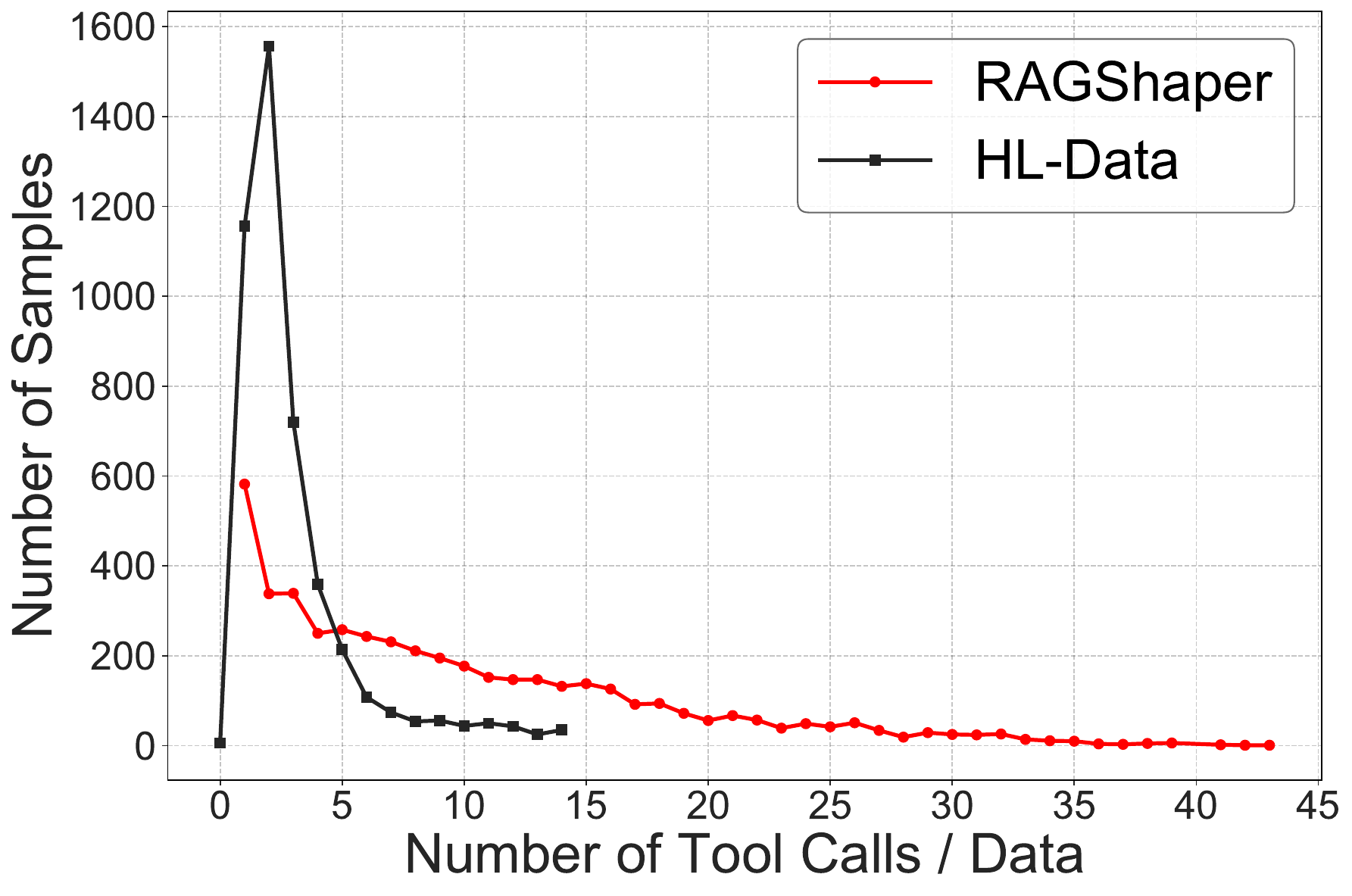}
   % \vspace{-3mm}
   \caption{Tool call statistics on 4.5k data on both \rsp~and HL-Data.}
   \label{fig: tool_performance}
   % \vspace{-1mm}
\end{figure}

\subsection{Trajectory Complexity 
% and Depth
Analysis}
\label{sec: trajectory_analysis}

To further investigate the reasoning quality of our synthesized corpora, we analyze the distribution of tool usage steps per trajectory. In Figure~\ref{fig: tool_performance}, we compare the trajectory depth of \rsp~against the human-labeled baseline (HL-Data).

\noindent\textbf{\rsp~synthesizes deeper, more complex reasoning tasks.} The distribution reveals a significant distinction in task complexity. HL-Data exhibits a sharp peak at 2-3 steps with a short tail, indicating that most human-annotated samples represent relatively shallow, few-shot reasoning tasks. In contrast, \rsp~presents a much broader, long-tailed distribution, with a substantial portion of trajectories requiring over 10, and up to 40+, steps. This confirms that our method successfully synthesizes tasks of higher difficulty.

\noindent\textbf{Longer trajectories encode richer agentic behaviors.} Crucially, a higher number of tool calls implies a richer density of agentic behaviors. The long-tail trajectories in \rsp~capture complex cognitive processes, such as navigating dead ends, verifying distractors, and performing extensive multi-hop planning, that are rarely present in the concise HL-Data. Furthermore, unlike generic datasets where models might answer from parametric memory, our distribution starts strictly after zero, ensuring that every trajectory involves necessary retrieval actions. This eliminates trivial ``direct answer'' cases and enforces a rigorous evidence-seeking process.

\begin{figure}
  \centering
    \setlength{\belowcaptionskip}{-3mm}
\includegraphics[width=\linewidth]{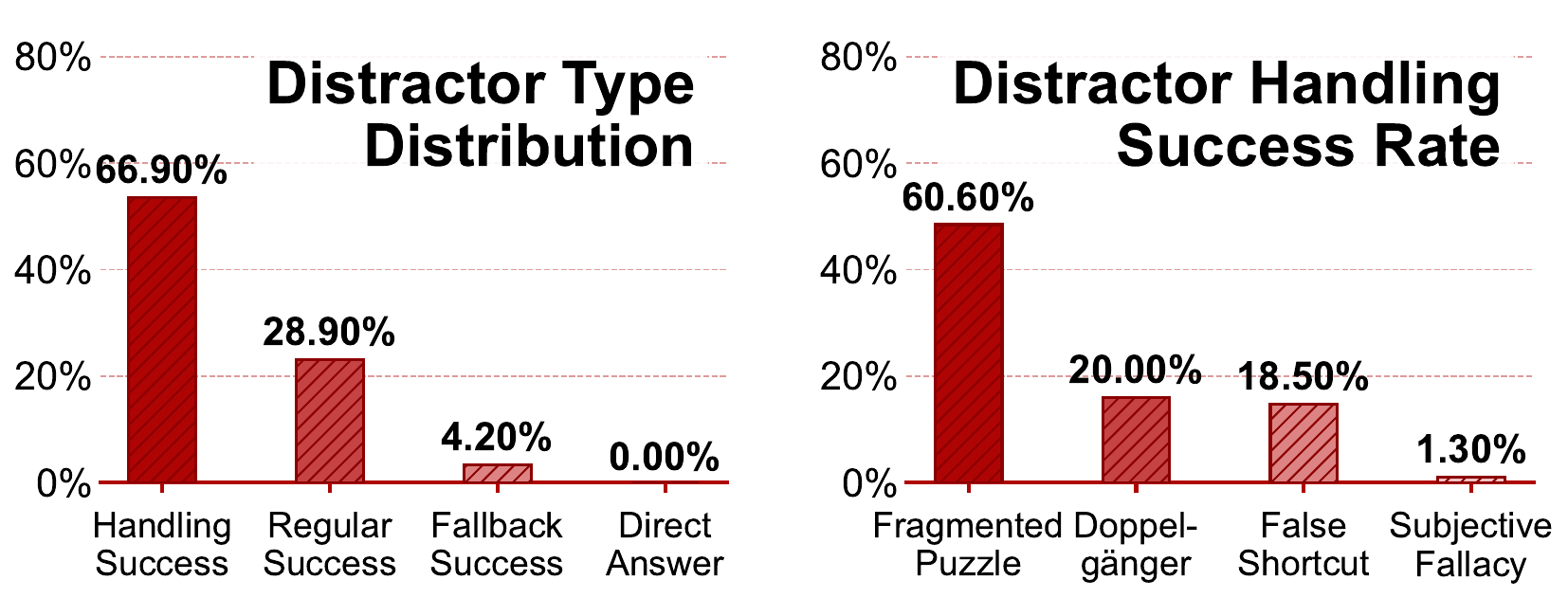}
   % \vspace{-3mm}
   \caption{Trajectory analysis.}
   \label{fig: traj_analysis}
   % \vspace{-5mm}
\end{figure}

\subsection{Trajectory Behavior Analysis}
\label{sec: traj_behavior}

To understand the underlying mechanisms of our model's success, we analyze the distribution of agent behaviors within the synthesized trajectories. We use LLM to tag each trajectory types, results are as visualized in Figure~\ref{fig: traj_analysis} (Left).

\noindent\textbf{Agents rely on rigorous retrieval rather than internal knowledge.} The analysis reveals that the majority of trajectories (66.90\%) are categorized as \textit{Handling Success}, where the agent successfully identifies and resolves the injected distractors to reach the correct answer. This high proportion, when viewed in conjunction with the high number of tool calls observed in Section~\ref{sec: trajectory_analysis}, confirms that our dataset is rich in high-quality agentic behaviors. The agent is not merely retrieving; it is actively reasoning against noise. Furthermore, the results indicate a strict reliance on retrieval rather than internal knowledge. The \textit{Direct Answer} rate is 0.00\%, and \textit{Fallback Success} (answering correctly despite failing to retrieve useful information) comprises only 4.20\%. This low prevalence of non-retrieval based answers demonstrates that the performance improvements are driven by the agent's enhanced ability to interact with external corpora, rather than by internal knowledge hallucinations or simple memorization.

\noindent\textbf{Complex cognitive traps provide headroom for future improvement.} Figure~\ref{fig: traj_analysis} (Right) further dissects the success rates across different distractor types tagged by LLM, revealing a distinct hierarchy of difficulty. While the agent shows competence in solving \textit{Fragmented Puzzles} (60.60\%), which primarily tests information aggregation, it encounters significant challenges with deeper cognitive traps. The low success rates for \textit{False Shortcut} (18.50\%) and the extremely challenging \textit{Subjective Fallacy} (1.30\%) suggest that the upper bound of our data's difficulty has \textbf{not} yet been reached. 
This ``headroom'' indicates that \rsp~provides a sufficiently complex environment for further research.
Future work could leverage this unexploited complexity through advanced training paradigms, such as reinforcement learning, to enable agents to master these subtle and adversarial reasoning scenarios.

\begin{figure*}[t]
  \centering
  \includegraphics[width=\linewidth]{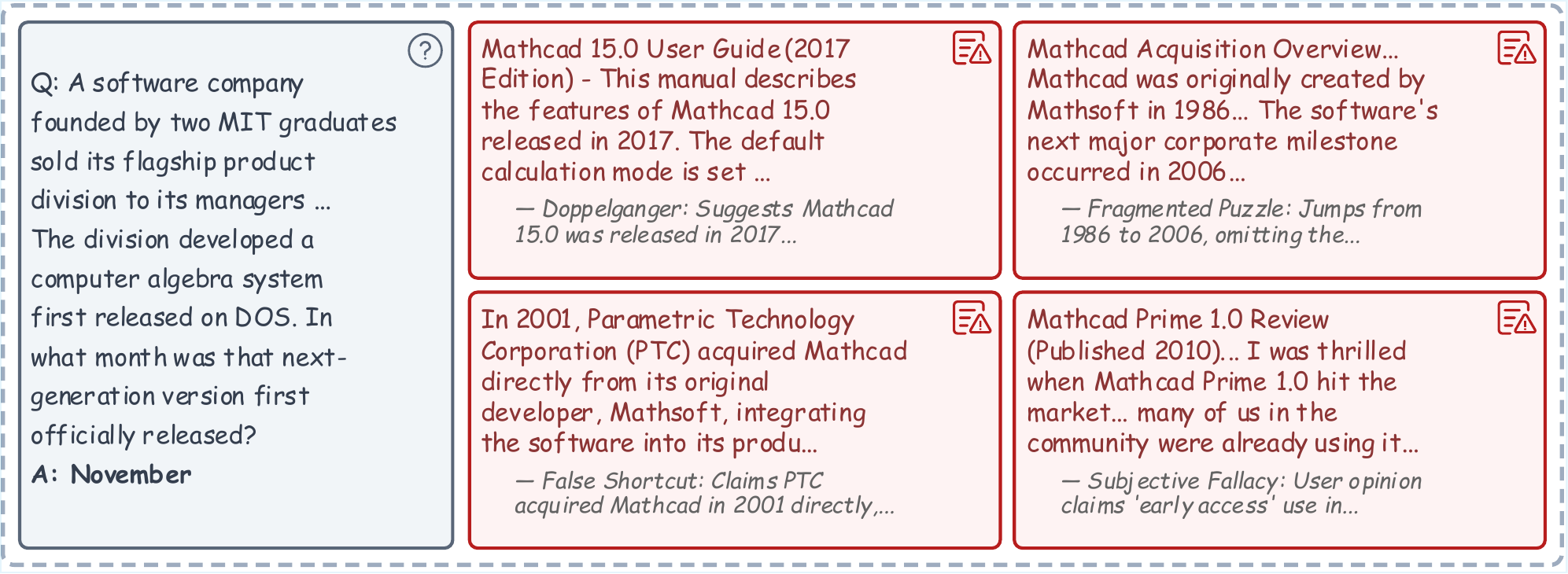}
  \caption{An illustrative example of the distractor taxonomy used in data synthesis. The figure visualizes four distinct categories of cognitive traps (Doppelganger, Fragmented Puzzle, False Shortcut, and Subjective Fallacy) designed to challenge the agent's retrieval and reasoning robustness.}
  \label{fig:qa_distractors}
\end{figure*}

% \vspace{-1mm}
\subsection{Generalization on Different Backbones}
\label{sec:generalization}

To further verify that the effectiveness of \rsp~is not limited to a specific model architecture, we extended our evaluation to a different backbone: Qwen3-4B-Think. 
We compare the performance of models fine-tuned on our synthesized data against those trained on HL-Data of the same scale (4.5k). 
The results are summarized in Table~\ref{tab: ablation_results}.

\noindent\textbf{\rsp~demonstrates strong generalization across diverse backbones.} 
As observed in the table, \rsp~consistently outperforms the HL-Data baseline, achieving a significant improvement in the overall average score. This confirms that the high-quality reasoning trajectories generated by our pipeline are universally beneficial and transferrable, rather than being overfitted to the specific characteristics of the certain experimental model.

% \vspace{-1mm}
\subsection{Case Study}
\label{sec: case_study}

Figure~\ref{fig:qa_distractors} shows a QA case with its distracting documents. We add reasons why these distracting documents can elicit sophisticated behaviours. Our method can generate various and effective distractors to stimulate advanced abilities of RAG agent.

\section{Related Work}
\label{sec: related_work}
\vspace{-1mm}
% \subsection{Retrieval-Augmented Generation}

% \noindent\paragraph{Prompt-based Approaches.}
% These methods enhance RAG at inference time without updating parameters. Strategies include interleaving retrieval with chain-of-thought reasoning \cite{shao2023enhancing, trivedi2023interleaving} and active retrieval mechanisms that trigger search based on generation confidence \cite{jiang2023active}. Other approaches focus on context compression to optimize information flow \cite{li2023unlocking, jiang2024longllmlingua, xu2023recomp, leeagent}. Recently, proprietary models like Search-o1 \cite{li2025search,sunoda} have achieved competitive results by integrating retrieval tools directly into reasoning.

% \noindent\paragraph{Learning-based Approaches.}
% To further improve performance, researchers train agents to synergize retrieval and generation. This involves modeling the process as a Markov Decision Process (MDP) \cite{guan2025deeprag, huang2025reinforced} or employing process-supervised reinforcement learning to optimize fine-grained steps \cite{zhang2025process, leng2025decex}. Additionally, strong open-weights models such as Search-R1 \cite{jin2025search} have emerged, equipping reasoning backbones with trainable search capabilities.
\noindent\paragraph{Retrieval-Augmented Reasoning Methods.}
Existing work improves RAG through both prompt-based and learning-based approaches. Prompt-based methods enhance inference without updating model parameters, including interleaving retrieval with chain-of-thought reasoning \cite{shao2023enhancing, trivedi2023interleaving}, triggering retrieval adaptively based on generation confidence \cite{jiang2023active}, and compressing context to improve information efficiency \cite{li2023unlocking, jiang2024longllmlingua, xu2023recomp, leeagent}. More recently, proprietary systems such as Search-o1 integrate retrieval tools directly into the reasoning process and achieve competitive performance \cite{li2025search, sunoda}. Learning-based approaches further improve performance by training agents to coordinate retrieval and generation, often formulating the process as a Markov Decision Process \cite{guan2025deeprag, huang2025reinforced} or applying process-supervised reinforcement learning for fine-grained optimization \cite{zhang2025process, leng2025decex}. In addition, strong open-weight models like Search-R1 equip reasoning backbones with trainable search capabilities \cite{jin2025search}.

% \vspace{-1mm}
\noindent\paragraph{Data for RAG.}
High-quality RAG systems often rely on human-labeled supervision. Standard baselines~\cite{jin2025search, leng2025decex, yu2024rankrag, li2025r3} utilize datasets like HotpotQA and 2WikiMultiHopQA~\cite{yang2018hotpotqa, ho2020constructing}, which are manually curated to test multi-hop reasoning. However, constructing such datasets requires labor-intensive annotation to verify evidence chains, making them expensive and difficult to scale for general-purpose training.

% \vspace{-1mm}
\noindent\paragraph{Agentically Data Synthesis.} To address the data scarcity in training generalist agents, recent studies have pivoted towards agentic data synthesis, where agents are employed to generate high-quality training samples~\cite{gao2025beyond,chen2025agentfrontier,zhai2025agentevolver}. Auto-Explorer~\cite{guo2025auto} introduces an explorer agent that autonomously navigates and parses GUI environments to collect diverse state-action pairs without human intervention. Similarly, OS-Genesis~\cite{sun2025genesis} proposes a reverse task synthesis pipeline, where agents first interact with the environment to create trajectories, which are then retrospectively aligned with synthesized high-level instructions. For search agents, WebShaper~\cite{tao2025webshaper} utilizes a formalization-driven framework with an agentic Expander to iteratively generate complex queries and reasoning paths. Furthermore, DeepSeek-V3.2~\cite{liu2025deepseek} implements a large-scale synthesis pipeline, deploying specialized agents to construct and verify tasks across various domains, enhancing agent generalization.

% \vspace{-2mm}
\section{Conclusion}
\label{sec: conclusion}
% \vspace{-2mm}

We presented \rsp, a framework designed to overcome the scalability and quality limitations of human annotation for Agentic RAG. By leveraging the \texttt{InfoCurator}, we automate the construction of dense retrieval environments populated with adversarial distractors across \textit{Perception} and \textit{Cognition} dimensions. Furthermore, our constrained navigation strategy effectively captures robust error-correction behaviors from teacher agents. Empirical results demonstrate that models trained on our synthesized corpus significantly outperform baselines in complex settings. 

\section*{Limitations}
In this work, we leverage RAGShaper to construct sophisticated behaviours of RAG agent. However, as discussed in Section~\ref{sec: traj_behavior}, our data has not fully unlocked its potential. In future work, more adavanced approaches can be applied to our data with further training mechanisms.

\section*{Ethical Considerations}
This work uses publicly available wikipedia documents and entities. It won't contain any information that names or uniquely identifies individual people or offensive content. We only use AI for writing assistant.

\bibliography{custom}

\clearpage
\appendix

\section{Training Details}
\label{app: training_details}

We fine-tune the Qwen3-30B-A3B-Think\footnote{https://huggingface.co/Qwen/Qwen3-30B-A3B-Thinking-2507} and Qwen3-4B-Think\footnote{https://huggingface.co/Qwen/Qwen3-4B-Thinking-2507} models using the Megatron-LM framework. We extend the context length to 128k. We employ the AdamW optimizer with a precision-aware configuration, coupled with a cosine decay learning rate scheduler. This scheduler features a peak learning rate of $1.0 \times 10^{-5}$, a minimum learning rate of $1.0 \times 10^{-6}$, and a 5\% warmup phase. The global batch sizes are configured as 16 for Qwen3-30B-A3B-Think and 40 for Qwen3-4B-Think. Both models are trained for 5 epochs, and the checkpoint exhibiting the best performance is selected for evaluation.

\subsection{Evaluation Metrics}
Following standard open-domain Question Answering protocols, we employ two primary metrics:
\begin{itemize}
    \item \textbf{Exact Match (EM):} Measures the percentage of predictions that match one of the ground-truth answers exactly after normalization.
    \item \textbf{F1 Score:} Measures the token overlap between the predicted answer and the ground truth, providing a granular assessment of partial correctness.
\end{itemize}

\section{Evaluation Benchmarks}
\label{app: eval_benchmark}
We utilize four datasets to evaluate distinct aspects of retrieval and reasoning:
\begin{itemize}
    \item \textbf{Natural Questions (NQ) \cite{kwiatkowski2019natural}:} A large-scale benchmark comprising real user queries issued to Google Search. We utilize the open-domain split, requiring the agent to retrieve answers from the entire Wikipedia corpus.
    \item \textbf{PopQA \cite{mallen2023popqa}:} Designed to evaluate factual retrieval for long-tail entities. This dataset contains queries where parametric memory typically fails, thereby necessitating precise external retrieval.
    \item \textbf{AmbigQA \cite{min2020ambigqa}:} Derived from NQ, this dataset focuses on ambiguous queries with multiple plausible answers. It challenges the agent's ability to disambiguate user intent and navigate noisy retrieval contexts.
    \item \textbf{Bamboogle \cite{press2023measuring}:} A "google-proof" dataset crafted to test multi-hop reasoning. Questions require synthesizing information from multiple distinct documents rather than locating a single direct answer.
\end{itemize}

\section{Baseline Details} % "Baselines Details" -> "Baseline Details"
\label{app: baseline_details}
We compare our approach against the following competitive baselines:

\paragraph{Prompt-Based Methods.} These utilize fixed LLMs with advanced prompting or retrieval strategies:
\begin{itemize}
    \item \textbf{Iter-RetGen \cite{shao2023enhancing}:} Iteratively synergizes retrieval and generation, utilizing model outputs to refine subsequent retrieval queries.
    \item \textbf{IR-CoT \cite{trivedi2023interleaving}:} Interleaves chain-of-thought reasoning with retrieval steps to guide multi-hop question answering.
    \item \textbf{FLARE \cite{jiang2023active}:} An active retrieval strategy that triggers information seeking only when the model generates low-confidence tokens.
    \item \textbf{Context Optimization Methods:} Including \textbf{Selective-Context \cite{li2023unlocking}}, \textbf{LongLLMLingua \cite{jiang2024longllmlingua}}, and \textbf{RECOMP \cite{xu2023recomp}}, which focus on compressing and selecting context to optimize information flow to the generator.
    \item \textbf{Search-o1 \cite{li2025search}:} A proprietary baseline utilizing the OpenAI o1-preview model equipped with search tools, representing state-of-the-art inference-time reasoning.
\end{itemize}

\paragraph{Learning-Based Methods.} These involve training the agent or retriever to enhance performance:
\begin{itemize}
    \item \textbf{DeepRAG \cite{guan2025deeprag}:} Models retrieval-augmented reasoning as a Markov Decision Process (MDP) for adaptive retrieval.
    \item \textbf{IKEA \cite{huang2025reinforced}:} A reinforced agent designed to synergize internal parametric knowledge with external search, optimizing for efficiency.
    \item \textbf{ReasonRAG \cite{zhang2025process}:} Utilizes process-supervised reinforcement learning with fine-grained rewards for query and answer generation.
    \item \textbf{DecEx-RAG \cite{leng2025decex}:} Enhances agentic RAG via decision and execution optimization using process supervision.
    \item \textbf{Search-R1 \cite{jin2025search}:} Utilizes the DeepSeek-R1 model equipped with search capabilities, serving as a representative strong open-weights reasoning model.
    \item \textbf{HL-Data:} A supervised baseline fine-tuned on high-quality human-labeled datasets (combining HotpotQA and 2WikiMultiHopQA)~\cite{yang2018hotpotqa, ho2020constructing}. This matches the scale of our synthesized data to serve as a control for data quality.
\end{itemize}

\begin{figure}[t]
  \centering
\includegraphics[width=0.78\linewidth]{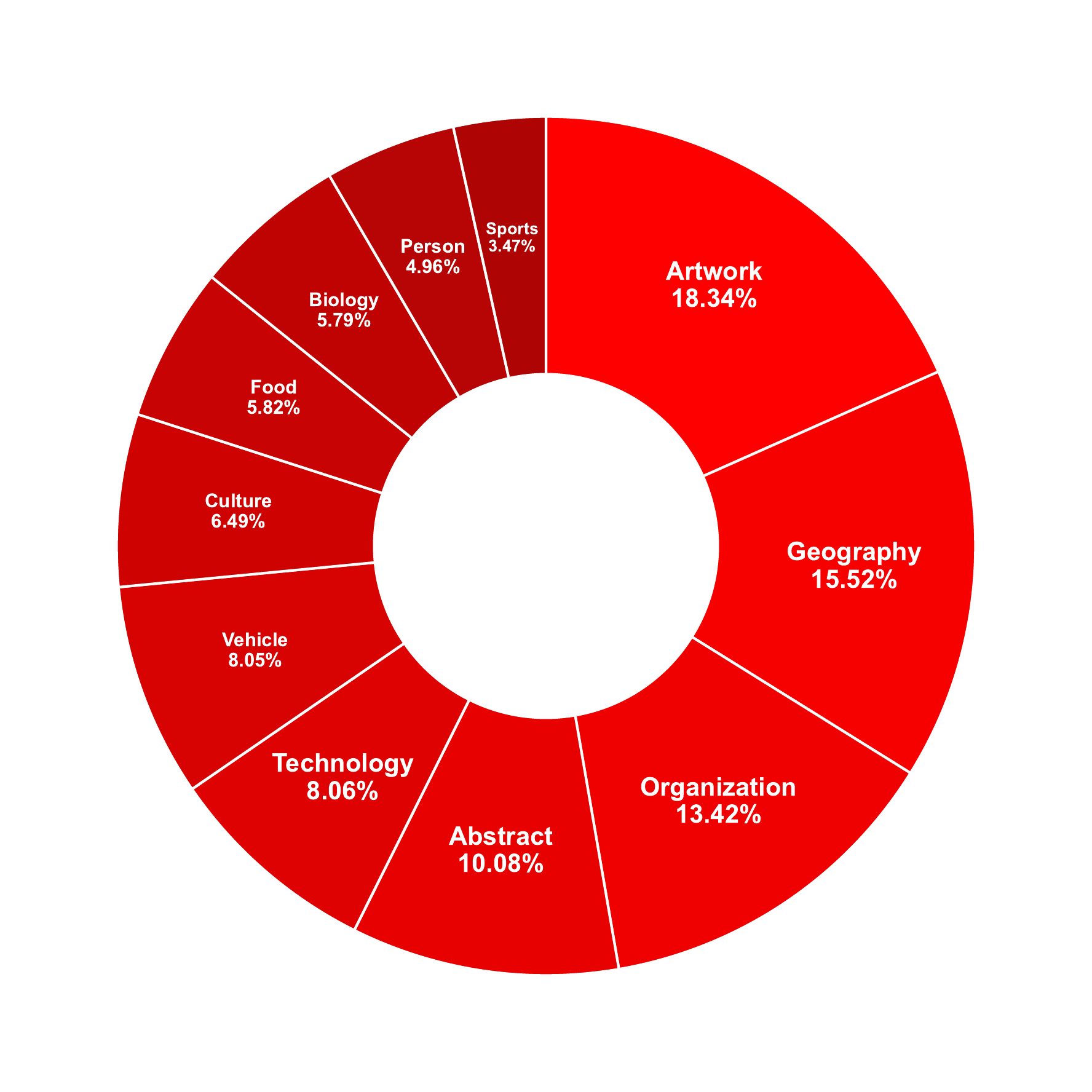}
   \vspace{-5mm}
   \caption{Domain distribution.}
   \label{fig: domain_dist}
   \vspace{-5mm}
\end{figure}

\section{Deployment and Inference Details}
We deployed gpt-oss-120b\footnote{https://huggingface.co/openai/gpt-oss-120b} and our trained models using the vLLM inference engine on 8$\times$H20 GPUs. For gpt-oss-120b, we set the maximum context length to $100,000$ tokens. The tool-call parser was configured to use the \texttt{openai} format. For our trained models, we utilized the \texttt{hermes} tool-call parser. All models were served via OpenAI-compatible APIs to maintain a consistent interface. We used FAISS~\cite{johnson2019billion} to support fast similarity search.

\section{Domain Diversity Analysis}
\label{sec: domain_analysis}

To verify the semantic coverage and generalization potential of our synthesized corpus, we conducted a domain analysis of the generated tasks. We employed an LLM to automatically classify the topic of each question-answer pair within the dataset.

\noindent\textbf{RAGShaper achieves exceptional and balanced domain diversity.} The resulting distribution, illustrated in Figure~\ref{fig: domain_dist}, demonstrates that \rsp~achieves exceptional diversity. The synthesized tasks span a broad spectrum of knowledge areas, ranging from humanities and social sciences (e.g., \textit{Artwork}, \textit{Culture}, \textit{Organization}) to STEM fields (e.g., \textit{Technology}, \textit{Biology}). Notably, the distribution is remarkably balanced; no single domain dominates the corpus, with the largest category (\textit{Artwork}) accounting for only 18.34\% and substantial representation even in complex categories like \textit{Abstract} (10.08\%). This structural uniformity ensures that the trained agent avoids overfitting to specific entity types and instead acquires generalized retrieval and reasoning skills applicable across diverse real-world contexts.

% === 建议将以下颜色定义移至导言区 (Before \begin{document}) ===
\definecolor{TrajFrame}{HTML}{AF0404}
\definecolor{TrajBack}{HTML}{F5F5F5}
\definecolor{WrongBg}{HTML}{F8D7DA}
\definecolor{WrongText}{HTML}{721C24}
\definecolor{CorrectBg}{HTML}{D4EDDA}
\definecolor{CorrectText}{HTML}{155724}
% ==========================================================

\begin{figure*}[p]
    \centering
    
    \begin{tcolorbox}[
        title={\textbf{Trajectory Example: Handling Doppelgänger Entities}},
        fontupper=\small, 
        colback=TrajBack!20,
        colframe=TrajFrame!80,
        coltitle=white,
        width=\linewidth,
        arc=3pt, boxrule=1.5pt,
        boxsep=4pt,
        left=4pt, right=4pt, top=4pt, bottom=4pt
    ]
        \setlength{\parskip}{0.5em}

        \textbf{User Question:} 
        A major literary work commissioned by the Holy Roman Emperor whose reign began in 1508 was part of his grand artistic legacy. While this patron commissioned famous manuscript anthologies during this period, this specific allegorical epic was distinctively designed for the printing press to ensure a wider audience. What is the exact publication year of its first edition?

        \rule{\linewidth}{0.5pt} 

        \textbf{[Thought 1]} 
        I need to identify the work and its date based on the constraints...
        I will search for works commissioned by Maximilian I during this period...

        \textbf{[Action 1]} 
        \texttt{Dense\_Search(query="Maximilian I commissioned literary works completed 1508-1519 manuscript vs print")}
        
        \textbf{[Observation 1]}

        % --- 作者标注区域：干扰项 ---
        \begin{tcolorbox}[colback=WrongBg, frame hidden, arc=2pt, boxrule=0pt, left=3pt, right=3pt, top=2pt, bottom=2pt, sharp corners]
            \textcolor{WrongText}{\textbf{\textit{Candidate A (Source 1):}}}
            \newline
            % [Fixed Quotes]
            \textcolor{WrongText}{``The \textbf{Ambraser Heldenbuch}... written by Hans Ried from \textbf{1504 to 1516}. It is a splendid \textbf{manuscript}...''}
            \newline
            % [Analysis] 标签明确为作者注释
            \textbf{\textcolor{WrongText}{[Analysis: Potential Doppelgänger]}} \textcolor{WrongText}{Completion (1516) fits the reign. However, the source identifies it explicitly as a ``manuscript,'' which conflicts with the print constraint.}
        \end{tcolorbox}
        
        % --- 作者标注区域：正确项 ---
        \begin{tcolorbox}[colback=CorrectBg, frame hidden, arc=2pt, boxrule=0pt, left=3pt, right=3pt, top=2pt, bottom=2pt, sharp corners, code={\vspace{1mm}}]
            \textcolor{CorrectText}{\textbf{\textit{Candidate B (Source 2):}}}
            \newline
            % [Fixed Quotes]
            \textcolor{CorrectText}{``The allegorical epic \textbf{Theuerdank}... The first edition was \textbf{printed} in Nuremberg and \textbf{published in 1517}...''}
            \newline
            % [Analysis] 标签明确为作者注释
            \textbf{\textcolor{CorrectText}{[Analysis]}} \textcolor{CorrectText}{Publication (1517) fits the reign. The format (printed) matches the constraint.}
        \end{tcolorbox}

        \textbf{[Thought 2]} 
        The retrieved documents mention two major works associated with Maximilian I during this period: the \textit{Ambraser Heldenbuch} and \textit{Theuerdank}.
        
        I need to strictly apply the format constraint. The snippet regarding the \textit{Ambraser Heldenbuch} explicitly describes it as a ``manuscript,'' which contradicts the user's request for a ``publication year'' (implying a printed edition). 
        
        In contrast, the description of \textit{Theuerdank} states it was ``printed'' and ``published in 1517.'' This aligns with all constraints.

        \textbf{[Action 2]} 
        \texttt{Dense\_Search(query="Theuerdank first edition publication year")}

        \textbf{[Observation 2]}
        \textit{Source 3:} ``...The \textbf{Theuerdank (1517)} was the first of his massive printed projects...''

        \textbf{[Thought 3]} 
        The evidence is conclusive...
        1. Patron: Maximilian I. 2. Date: 1517. 3. Format: Printed.

        \textbf{[Final Answer]} 
        1517
    \end{tcolorbox}
    
    % === Caption ===
    \caption{Detailed trajectory of the ``Doppelgänger Handling'' scenario. The agent successfully disambiguates between two contemporaneous works commissioned by Maximilian I. \textbf{Note: The colored blocks marked `[Analysis]' are authorial annotations added to illustrate the critical distinction (Manuscript vs. Print) and do not represent the agent's raw observation.}}
    \label{fig:traj_example_doppelganger}
\end{figure*}

\begin{figure*}[p]
    \centering
    
    % 如果颜色已经在前面定义过，这里不需要重复定义。
    % 仅保留 tcolorbox 设置
    \begin{tcolorbox}[
        title={\textbf{Trajectory Example: Countering Subjective Fallacy}},
        fontupper=\small, 
        colback=TrajBack!20,
        colframe=TrajFrame!80,
        coltitle=white,
        width=\linewidth,
        arc=3pt, boxrule=1.5pt,
        boxsep=4pt,
        left=4pt, right=4pt, top=4pt, bottom=4pt
    ]
        \setlength{\parskip}{0.5em}

        \textbf{User Question:} 
        An unofficial proposal to name a synthetic element after Humphry Davy circulated in the scientific community in 2014. The element's provisional IUPAC name was `ununpentium', its discovery was recognized in December 2015, and its official name was approved on a specific date in November 2016 after the discoverers exercised their naming rights. What is the official name of this element?

        \rule{\linewidth}{0.5pt} 

        \textbf{[Thought 1]} 
        I need to identify the official name of the element provisionally known as `ununpentium' (Element 115). 
        Key constraints:
        1. Recognized in Dec 2015.
        2. Official name approved in November 2016.
        3. A proposal for ``Davy'' existed, but I need to confirm if it was the one approved or just a rumor.
        
        Initial knowledge suggests Element 115 is ``Moscovium'', but the mention of ``Humphry Davy'' requires verification to ensure I don't fall for a trick question about a rejected or popular unofficial proposal.

        \textbf{[Action 1]} 
        \texttt{Dense\_Search(query="Moscovium name approved November 2016 official name element 115")}
        
        \textbf{[Observation 1]}

        % 正确信息
        \begin{tcolorbox}[colback=white, frame hidden, arc=2pt, boxrule=0pt, left=3pt, right=3pt, top=2pt, bottom=2pt, sharp corners, borderline west={2pt}{0pt}{gray}]
            % [Fixed Quotes]
            \textit{Source 2 \& 5:} ``Moscovium... On 28 November 2016, it was officially named after the Moscow Oblast...'' / ``In June 2016, IUPAC endorsed the proposal... formally accepted on 28 November 2016.''
        \end{tcolorbox}

        % 错误信息高亮
        \begin{tcolorbox}[colback=WrongBg, frame hidden, arc=2pt, boxrule=0pt, left=3pt, right=3pt, top=2pt, bottom=2pt, sharp corners]
            \textcolor{WrongText}{\textbf{\textit{Source 3 (Conflicting Data):}}}
            \newline
            % [Fixed Quotes]
            \textcolor{WrongText}{``The chemistry community finally recognized Sir Humphry Davy's legacy when the IUPAC officially approved the name \textbf{`Davyum'} for element 115 in 2016...''}
            \newline
            \textbf{\textcolor{WrongText}{[Analysis: Potential Fallacy]}} \textcolor{WrongText}{This snippet claims ``Davyum'' was approved. This directly contradicts Sources 2 and 5 which cite ``Moscovium''.}
        \end{tcolorbox}

        \textbf{[Thought 2]} 
        I have a contradiction. Source 3 claims ``Davyum'' was approved, while Sources 2 and 5 claim ``Moscovium''. 
        The prompt mentions an ``unofficial proposal'' for Davy in 2014, which likely explains the existence of the ``Davyum'' rumor. However, the official approval in November 2016 is widely documented as Moscovium.
        I need to verify the origin of the ``Davy'' proposal vs the naming rights of the discoverers (likely Russian/American).

        \textbf{[Action 2]} 
        \texttt{Dense\_Search(query="origin of Davyum proposal element 115 vs Moscovium naming rights")}

        \textbf{[Observation 2]}
        \textit
        ...Dubna team (Russia) proposed `moscovium' honoring the Moscow region... The `Davyum' name was suggested by an online petition/campaign but was not the discoverers' choice... On 30 December 2015, discovery recognized...
        (The search results confirm Moscovium is the official choice by the discoverers).

        \textbf{[Thought 3]} 
        I can now resolve the conflict. 
        
        % 纠正逻辑高亮
        \begin{tcolorbox}[colback=CorrectBg, frame hidden, arc=2pt, boxrule=0pt, left=3pt, right=3pt, top=2pt, bottom=2pt, sharp corners]
            \textcolor{CorrectText}{\textbf{[Critical Reasoning]}}
            \newline
            % [Fixed Quotes]
            \textcolor{CorrectText}{The snippet from Source 3 regarding `Davyum' appears to be \textbf{spurious misinformation} based on unofficial campaigns. It contradicts the specific dates (28 Nov 2016) associated with Moscovium in multiple other reliable sources.}
            \vspace{1mm}
            \newline
            \textcolor{CorrectText}{The question mentions a proposal for Davy, but the official name approved on the specific date (Nov 2016) after discoverers exercised naming rights is \textbf{``Moscovium''}.}
        \end{tcolorbox}

        \vspace{2mm}
        \textbf{[Final Answer]} 
        Moscovium
    \end{tcolorbox}
    
    \caption{Trajectory demonstrating resilience against the ``Subjective Fallacy''. The retrieval system returned a \protect\colorbox{WrongBg}{\textcolor{WrongText}{hallucinated or fake document}} claiming Element 115 was named ``Davyum''. The agent cross-referenced this with other sources, correctly identified the distractor as misinformation derived from unofficial proposals, and reasoned using the logic highlighted in the \protect\colorbox{CorrectBg}{\textcolor{CorrectText}{green block}}, thereby demonstrating the system's \textbf{robustness against conflicting information} and hallucinated sources.}
    \label{fig:traj_example_fallacy}
\end{figure*}

\begin{figure*}[t] 
    \centering
    \includegraphics[width=\linewidth]{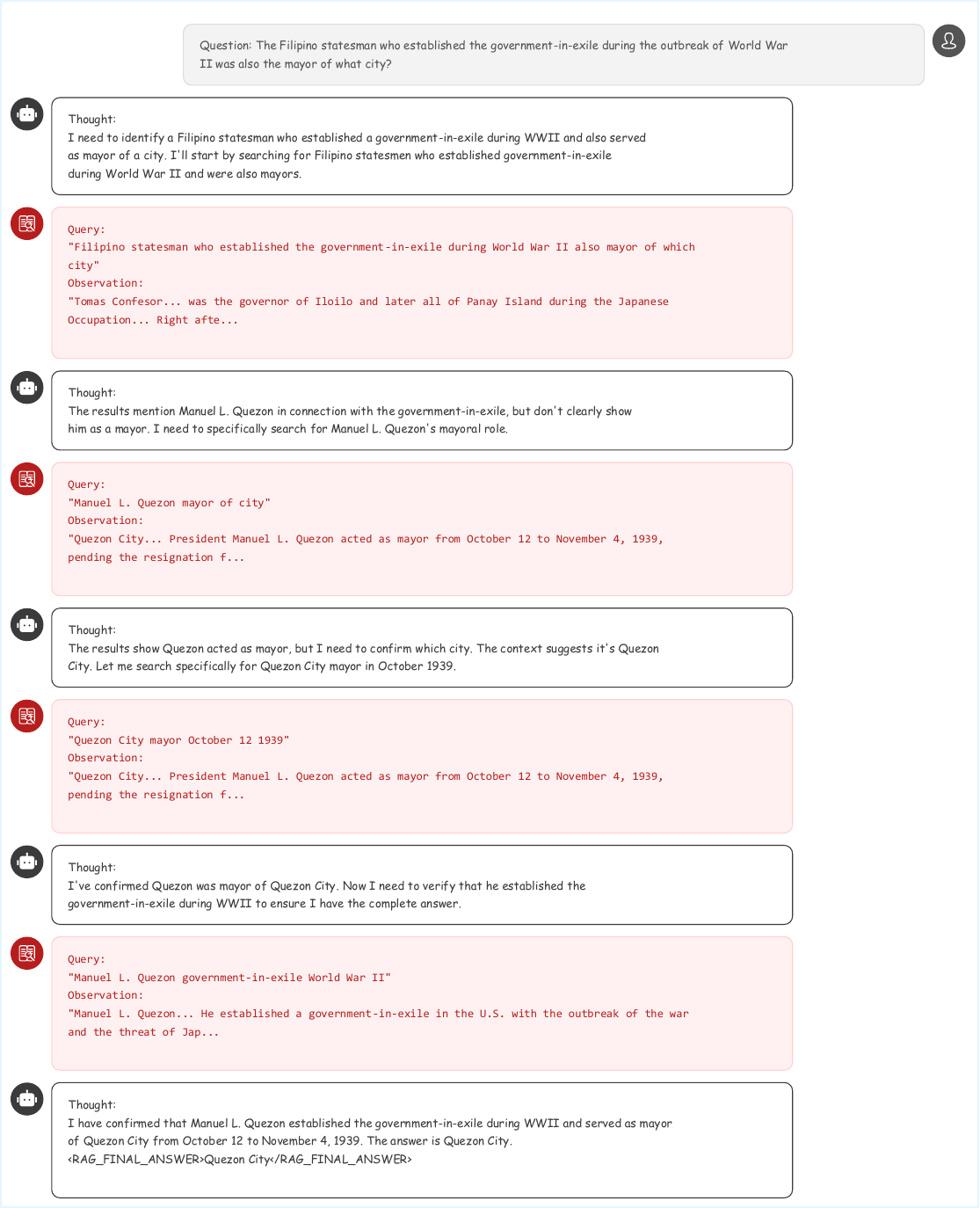}
    \caption{Trajectory generated by 4.5k RAGShaper trained on the Qwen3-30B-A3B-Thinking-2507. The figure shows the agent solving a multi-hop question: it first identifies the likely candidate (Manuel L. Quezon) from the "government-in-exile" clue, pivots to verify the specific mayoral role (Action 2), discovering he acted as mayor of \textbf{Quezon City}, and finally cross-references the WWII context (Action 3) to validate the answer.}
    \label{fig:traj_quezon_pdf}
\end{figure*}
\begin{figure*}
    \definecolor{solarizedBase}{RGB}{245, 245, 245}
    \definecolor{solarizedFrame}{RGB}{175, 4, 4}
    \definecolor{codeGreen}{RGB}{133, 153, 0}
    \definecolor{codeCyan}{RGB}{42, 161, 152}

    \lstset{
        basicstyle=\ttfamily\footnotesize,
        breaklines=true,
        columns=fullflexible,
        stringstyle=\color{codeCyan},
        keywordstyle=\color{codeGreen},
        upquote=true,
    }

    \begin{tcolorbox}[
        title={Tool Schema},
        colback=solarizedBase!20,
        colframe=solarizedFrame,
        coltitle=white,
        fonttitle=\bfseries,
        width=\textwidth
    ]
    \textbf{\textit{Dense Search}}
    \begin{lstlisting}
{
    "type": "function",
    "function": {
        "name": "query_knowledge_base_dense",
        "description": "[Dense Search] Semantic vector search over the knowledge base. Falls back to configured top_k or 5.",
        "parameters": {
            "type": "object",
            "properties": {
                "query": {
                    "type": "string",
                    "description": "Natural language question or statement to retrieve against the KB.",
                    "minLength": 1
                },
                "top_k": {
                    "type": "integer",
                    "description": "Override for number of results; positive integer.",
                    "minimum": 1
                }
            },
            "required": ["query"]
        }
    }
}
    \end{lstlisting}
    \end{tcolorbox}
    \caption{Tool schema definition for the dense vector retrieval tool (\textit{query\_knowledge\_base\_dense}).}
    \label{fig:dense_tool_schema}
\end{figure*}

\begin{figure*}[h]
    \centering
    % === 颜色定义 ===
    \definecolor{TrajFrame}{HTML}{AF0404}
    \definecolor{TrajBack}{HTML}{F5F5F5}
    \definecolor{CodeGreen}{RGB}{80, 120, 80}
    \definecolor{CodePurple}{RGB}{100, 50, 150}

    % === 设置代码样式 (虽然删除了大段JSON，但保留设置以防万一) ===
    \lstset{
        basicstyle=\ttfamily\scriptsize,
        breaklines=true,
        breakatwhitespace=true,
        columns=fullflexible,
        keepspaces=true,
        frame=none,
        tabsize=2
    }

    \begin{tcolorbox}[
        title={\textbf{Core Prompt of Exploration in Information Curation}},
        fontupper=\small,
        colback=TrajBack!20,
        colframe=TrajFrame!80,
        coltitle=white,
        width=\linewidth,
        arc=3pt, boxrule=1.5pt,
        boxsep=4pt,
        left=4pt, right=4pt, top=4pt, bottom=4pt
    ]

\textbf{=== PRIMARY GOAL ===}
Sample a trajectory that will later support a \textbf{LOW-ENTRANCE but DEEP multi-hop QA}.
You are not just collecting facts — you are building a dependency chain (A$\to$B$\to$C$\to$D...) plus confusable-but-disambiguatable negative documents.

\medskip
\textbf{=== SAMPLING STRATEGY \& RULES ===}

\textbf{1) Build a Multi-hop Backbone (Depth-First Chain)}
\begin{itemize}[leftmargin=*, nosep]
    \item Target $\ge$ 10 dependent hops whenever possible (A$\to$B$\to$C$\to$D...).
    \item Each retrieval step MUST unlock a \textbf{NEW} entity/relation needed for the NEXT hop.
    \item Do NOT get stuck circling the same entity. Revisit only to cross-check hard metadata.
\end{itemize}

\textbf{2) Pack Compact Evidence per Hop}
\begin{itemize}[leftmargin=*, nosep]
    \item Capture 1--2 short, quotable snippets per hop that clearly state the relation.
    \item Capture at least ONE \textbf{hard metadata item} (year/date/version/ID/count) that can be cross-checked.
    \item Ensure FINAL answer-critical metadata is supported by $\ge$ 2 independent observations.
\end{itemize}

\textbf{3) Generate Negative Docs Early \& Repeatedly}
\begin{itemize}[leftmargin=*, nosep]
    % 注意：下划线已转义，\ge 已放入数学模式
    \item \textbf{Tool Usage:} You must use \texttt{write\_distractor\_docs} (pass \texttt{distractor\_texts} list). Do NOT call an LLM for this; write the text yourself.
    \item \textbf{Timing:} Create negative docs after the FIRST successful retrieval and after each key hop (especially when hard metadata appears).
    \item \textbf{Quantity:} Min $\ge$ 3 calls total; total $\ge$ 5 distractor documents per seed. Diversify dimensions.
\end{itemize}

\textbf{4) Safety Rule: Disambiguation is Mandatory}
\begin{itemize}[leftmargin=*, nosep]
    \item The solver cannot know which doc is a distractor. Every negative doc MUST be logically distinguishable (e.g., specific year, version, or scope).
    \item \textit{Bad:} "Founded in 2015" vs "Founded in 2016" with no other context.
    \item \textit{Good:} "2015 Annual Report (Audited)" vs "2016 Preliminary Draft".
\end{itemize}

\medskip
\textbf{=== DIMENSION GUIDANCE (Types of Negative Docs) ===}
\begin{itemize}[leftmargin=*, nosep]
    \item \textbf{[A] Doppelganger:} Adjacent-edition doc (e.g., 2015 vs 2016 manual). Change one spec/value but keep the rest similar. Make the edition explicit.
    \item \textbf{[B] False Shortcut:} A doc claiming A$\to$C directly (skipping B) with hedged phrasing, contradicting the true A$\to$B$\to$C chain.
    \item \textbf{[C] Fragmented Puzzle:} Docs containing only a subset of information, looking locally plausible but incomplete.
    \item \textbf{[D] Subjective Fallacy:} Review/Opinion tone with one plausible factual objective error (e.g., wrong model number).
\end{itemize}

    \end{tcolorbox}
    \caption{The \textbf{Core Exploration Trajectory} prompt. Unlike standard retrieval, this prompt drives the agent to proactively construct deep dependency chains (10+ hops) and synthesize "Doppelgänger" or "False Shortcut" negative documents during the rollout, laying the groundwork for high-complexity puzzle generation.}
    \label{fig:core_exploration_prompt}
\end{figure*}
\begin{figure*}[p] % [p] 强制独占一页
    \centering
    % === 颜色定义 ===
    \definecolor{TrajFrame}{HTML}{AF0404}
    \definecolor{TrajBack}{HTML}{F5F5F5}
    
    % === 代码样式调整 ===
    \lstset{
        % --- 关键修改1：使用 footnotesize 平衡大小和空间 ---
        basicstyle=\ttfamily\footnotesize, 
        breaklines=true,
        breakatwhitespace=true,
        columns=fullflexible,
        keepspaces=true,
        frame=none,
        tabsize=2,
        showstringspaces=false,
        % --- 关键修改2：收紧垂直间距 ---
        % lineskip=2pt, % 去掉额外的行间距
        aboveskip=2pt,  % 减少代码块上方的间距
        belowskip=2pt   % 减少代码块下方的间距
    }

    \begin{tcolorbox}[
        title={\textbf{Prompt: Trajectory-to-QA Synthesis}},
        fontupper=\small, % 标题字体保持稍大
        colback=TrajBack!20,
        colframe=TrajFrame!80,
        coltitle=white,
        width=\linewidth,
        arc=3pt, boxrule=1.5pt,
        % --- 关键修改3：收紧盒子内部边距 ---
        boxsep=3pt, % 从 6pt 恢复到较小值
        left=4pt, right=4pt, top=4pt, bottom=4pt
    ]
    
Please synthesize a high-quality Q\&A pair based on the trajectory:\\

\#\# Question Requirements (Crucial for Reasoning \& Brevity):
- The target answer must be a specific fact (e.g., a name, a date, a location, a count, or a yes/no status).

- **DO NOT** ask "How", "Why", or "Describe" questions that require long textual explanations. 

- **Anti-shortcut**: The question MUST NOT contain the answer text, and MUST NOT directly state the asked attribute in a definitional clause.

- **Low-entrance, deep-reasoning**: Keep the question to <=2 sentences and a small number of top-level clues; depth should come from a multi-hop dependency chain, not a long list of trivia.

- **Deep multi-hop (required)**: The question must require >=3 dependent hops to solve (chain dependency only; no star-shaped checklist).

- **Negative-doc confusability (required)**: If the trajectory includes negative docs (e.g., generated via write\_distractor\_docs), craft the question so that a careless solver could be misled by at least one negative doc into a plausible wrong answer/path, while the correct answer is still supported by authoritative evidence in the trajectory.
- The question should be a natural, factual, and self-contained question (e.g., don't include "What did the agent find...", "what is in the trajectory...", "according to the trajectory...", ...). It must seem like it never undergos a trajectory exploration in previous step. And don't mention "search" or "search results", or things like them.\\

\#\# Answer Requirements (Crucial for Strict Length):
- **Extreme Brevity**: The answer MUST be **less than or equal to one sentence, and contain only one entity**, or ideally just a **short phrase** (e.g., "1985", "The Treaty of Versailles", "Increased by 5\%").

- **No Fluff**: Do not use filler words like "According to the documents..." or "The answer is...". Provide ONLY the final answer value.

- **Groundedness**: The specific fact must be strictly derived from the provided trajectory observations without mentioning the trajectory or observation.\\

\#\# Required Explanations (for dataset traceability; NOT part of the question text):

- reasoning\_steps: Provide $>=3$ short, dependent steps that solve the QA using ONLY the trajectory evidence.

- negative\_aspect: Explain how negative doc(s) could mislead and what disambiguation defeats them. Mention the distractor dimension when possible.

- disambiguation: How to disambiguate the misleading claim.

- distraction\_text: The text that is used to distract the solver.

Return JSON EXACTLY in this schema (do not add extra fields):
\begin{lstlisting}
{
  "question": "question text",
  "answer": "short phrase or single sentence",
  "reasoning_steps": [
    {"hop": 1, "fact": "intermediate fact", "evidence": "snippet", "output": "entity/metadata"},
    {"hop": 2, "fact": "intermediate fact", "evidence": "snippet", "output": "entity/metadata"},
    ...
    {"hop": n, "fact": "final derivation", "evidence": "snippet", "output": "answer"},
  ],
  "negative_aspect": [
    {"dimension": "doppelganger|false_shortcut|fragmented_puzzle|subjective_fallacy", "misleading_claim": "claim", "disambiguation": "method", "distraction_text": "text"}
  ]
}
\end{lstlisting}
    \end{tcolorbox}
    \caption{The \textbf{QA Synthesis} prompt. This prompt consumes the trajectory generated in the previous step. It enforces strict constraints to ensure the synthesized question is "low-entrance" (concise) yet "deep-reasoning" (requires traversing the full dependency chain), and explicitly validates the effectiveness of the negative documents.}
    \label{fig:qa_synthesis_prompt}
\end{figure*}

\begin{figure*}[h]
    \centering
    % === 确保颜色已定义 ===
    \definecolor{TrajFrame}{HTML}{AF0404}
    \definecolor{TrajBack}{HTML}{F5F5F5}

    % === 设置 Prompt 专用的代码样式 ===
    \lstset{
        basicstyle=\ttfamily\footnotesize,
        breaklines=true,
        breakatwhitespace=true,
        columns=fullflexible,
        keepspaces=true,
        frame=none,
        showstringspaces=false
    }

    \begin{tcolorbox}[
        % --- 修改点：将标题改为强调 Rollout 阶段 ---
        title={\textbf{Prompt for Trajectory Rollout}},
        fontupper=\small,
        colback=TrajBack!20,
        colframe=TrajFrame!80,
        coltitle=white,
        width=\linewidth,
        arc=3pt, boxrule=1.5pt,
        boxsep=4pt,
        left=4pt, right=4pt, top=4pt, bottom=4pt
    ]
You are a helpful assistant. You need to use tools to solve the problem.
You have access to a Dense Retrieval system (semantic/vector search). You MUST use the dense retrieval tool to answer and verify. \\

\#\# Core Capabilities

- **Semantic Understanding**: The system matches the *meaning* of your query, not just exact words.

- **Handling Paraphrasing**: It can find relevant content even if different terminology is used.\\

\#\# Query Formulation Strategy

1. **Be Descriptive**: Write natural language queries that fully describe what you are looking for.
   - *Bad*: "revenue 2023"
   - *Good*: "What was the total revenue of the company in the fiscal year 2023?"
   
2. **Context Matters**: Include necessary context in the query string, as the retriever processes independent queries.

3. **Iterative Refinement**:
   - If results are too broad: Add specific constraints to your query.
   - If results are irrelevant: Rephrase the query using synonyms or related concepts.\\
   
\#\# Execution Protocol

1. Break complex multi-hop questions into separate, simpler queries.

2. Verify the retrieved content matches the user's intent.

3. If after multiple attempts (>5) no relevant information is found, try rephrasing your queries with different approaches.\\

\#\# Internal Knowledge Fallback Mechanism

When you have attempted multiple retrieval queries over several rounds but still cannot find the answer in the knowledge base, you should use your internal knowledge to provide the best possible answer. This is a fallback mechanism to ensure you can still help the user even when the knowledge base doesn't contain the required information. When using internal knowledge, clearly indicate this in your reasoning and wrap your answer in the final answer tags.\\

\#\# Critical Requirements

1. **Reasoning-Tool Consistency**: If your reasoning mentions using a tool (e.g., "Let's search", "We need to use the dense retrieval tool"), you MUST generate the corresponding tool\_calls. Do not stop at reasoning alone.

2. **Action Follow-through**: If you decide to use a tool in your reasoning, you must follow through with the actual tool call. Empty content with reasoning about tool usage is NOT a valid final answer.\\

\#\# Answer Strategy

1. The final answer should only contain the short answer to the question (few words), avoiding unnecessary reasoning content in the final output string.

2. **MANDATORY**: You MUST wrap the final answer inside \{FINAL\_ANSWER\_START\} and \{FINAL\_ANSWER\_END\} tags. Never provide an answer without these tags. Every response that contains an answer must use these tags.

3. **Answer Quality Requirements**:
   - The answer must be a specific entity: a name, place, number, date, ID, or other concrete information.
   
   - **DO NOT** use common words like "and", "or", "the", "of", "in", "is", "was", "are", "were", "a", "an", "as", "for", "with", "from", "to", "on", "at", "by", "this", "that", "these", "those" as your final answer.
   
   - Common words, articles, prepositions, and conjunctions are NOT valid answers. The answer should be a meaningful entity or piece of information that directly answers the question.
   
   - If the retrieved information does not contain a clear answer, indicate that you cannot find the answer, but still wrap your response in the answer tags.
4. Keep any reasoning or explanation outside the \{FINAL\_ANSWER\_START\} and \{FINAL\_ANSWER\_END\} tags.

    \end{tcolorbox}
    % --- 修改点：Caption 中也同步使用了 rollout phase 的表述 ---
    \caption{The full prompt used during the \textbf{trajectory rollout} phase to guide the agent in generating training data. It explicitly instructs the model on query formulation strategies, fallback mechanisms, and the strict formatting required for the final answer.}
    \label{fig:trajectory_prompt}
\end{figure*}

\begin{figure*}[h]
    \centering
    % === 颜色定义 (保持风格一致) ===
    \definecolor{TrajFrame}{HTML}{AF0404}
    \definecolor{TrajBack}{HTML}{F5F5F5}

    % === 代码样式设置 ===
    \lstset{
        basicstyle=\ttfamily\footnotesize,
        breaklines=true,
        breakatwhitespace=true,
        columns=fullflexible,
        keepspaces=true,
        frame=none,
        showstringspaces=false
    }

    \begin{tcolorbox}[
        title={\textbf{Prompt for Evaluation}},
        fontupper=\small,
        colback=TrajBack!20,
        colframe=TrajFrame!80,
        coltitle=white,
        width=\linewidth,
        arc=3pt, boxrule=1.5pt,
        boxsep=4pt,
        left=4pt, right=4pt, top=4pt, bottom=4pt
    ]
You are a helpful assistant. You need to use tools to solve the problem.
You have access to a Dense Retrieval system (semantic/vector search). You MUST use the dense retrieval tool to answer and verify. Do not attempt to use sparse retrieval tools as they are not available.\\

\#\# Core Capabilities

- **Semantic Understanding**: The system matches the *meaning* of your query, not just exact words.

- **Handling Paraphrasing**: It can find relevant content even if different terminology is used.\\

\#\# Query Formulation Strategy

1. **Be Descriptive**: Write natural language queries that fully describe what you are looking for.
   - *Bad*: "revenue 2023"
   - *Good*: "What was the total revenue of the company in the fiscal year 2023?"
   
2. **Context Matters**: Include necessary context in the query string, as the retriever processes independent queries.

3. **Iterative Refinement**:
   - If results are too broad: Add specific constraints to your query.
   - If results are irrelevant: Rephrase the query using synonyms or related concepts.\\
   
\#\# Execution Protocol

1. Break complex multi-hop questions into separate, simpler queries.

2. Verify the retrieved content matches the user's intent.

3. If after multiple attempts ($>5$) no relevant information is found, admit that the information is missing from the knowledge base.\\

\#\# Answer Strategy

1. The final answer should only contain the short answer to the question (few words), avoiding unnecessary reasoning content in the final output string.

2. Wrap the final answer inside <RAG\_FINAL\_ANSWER> and </RAG\_FINAL\_ANSWER>, and keep any reasoning outside the tokens.\\

\#\# Available Tools

- query\_knowledge\_base\_dense: [Dense Search] Semantic vector search over the knowledge base. Falls back to configured top\_k or 5.
    \end{tcolorbox}
    \caption{The prompt utilized during the evaluation phase. Compared to the training prompt, this version instructs the model to prioritize honesty by admitting when information is missing from the knowledge base, rather than falling back to internal knowledge. It also specifies XML-style tags for the final answer extraction.}
    \label{fig:eval_prompt}
\end{figure*}

\end{document}